\pdfoutput=1
\documentclass{article}

\usepackage{tufalabs}

\tufasetup{
  title = {A Predictive Law for On-Policy Self-Distillation From World Feedback},
  short-title = {A Predictive Law for On-Policy Self-Distillation From World Feedback},
  authors = {Tommy He \quad Jerome Sieber\textsuperscript{*} \quad Matteo Saponati\textsuperscript{*}},
  affiliation = {Tufa Labs, Z\"urich, Switzerland},
  pdf-authors = {Tommy He, Jerome Sieber, Matteo Saponati},
  contact = {tommy@tufalabs.ai},
  code-url = {https://github.com/Tufalabs/opsd-predictive-law},
  author-note = {\textsuperscript{*}Equal contribution.},
  abstract = {Moving beyond simple scalar rewards toward richer world feedback is a natural path to more scalable RL post-training. On-policy self-distillation (OPSD) is a promising recent approach that uses arbitrary feedback as learning signal, yet its reliability compared to established methods, such as GRPO, remains unclear. We identify a strikingly consistent linear correlation between the initial student–self-teacher performance gap and the final performance improvement in OPSD. This relationship holds across context types and model families, providing a powerful predictive law for anticipating the outcome of an OPSD configuration without running the full training procedure. Interestingly, we show that this linear predictability holds with model scale, suggesting a potential basis for new empirical scaling laws on larger models with stronger in-context learning capabilities. In essence, our findings show that OPSD performance can be predicted and tuned before training, offering a principled way to incorporate world feedback as a first-class component of the post-training pipeline.}
}

\begin{document}

\maketufatitle
\begin{figure*}[ht!]
\begin{center}
\includegraphics[width=0.98\linewidth]{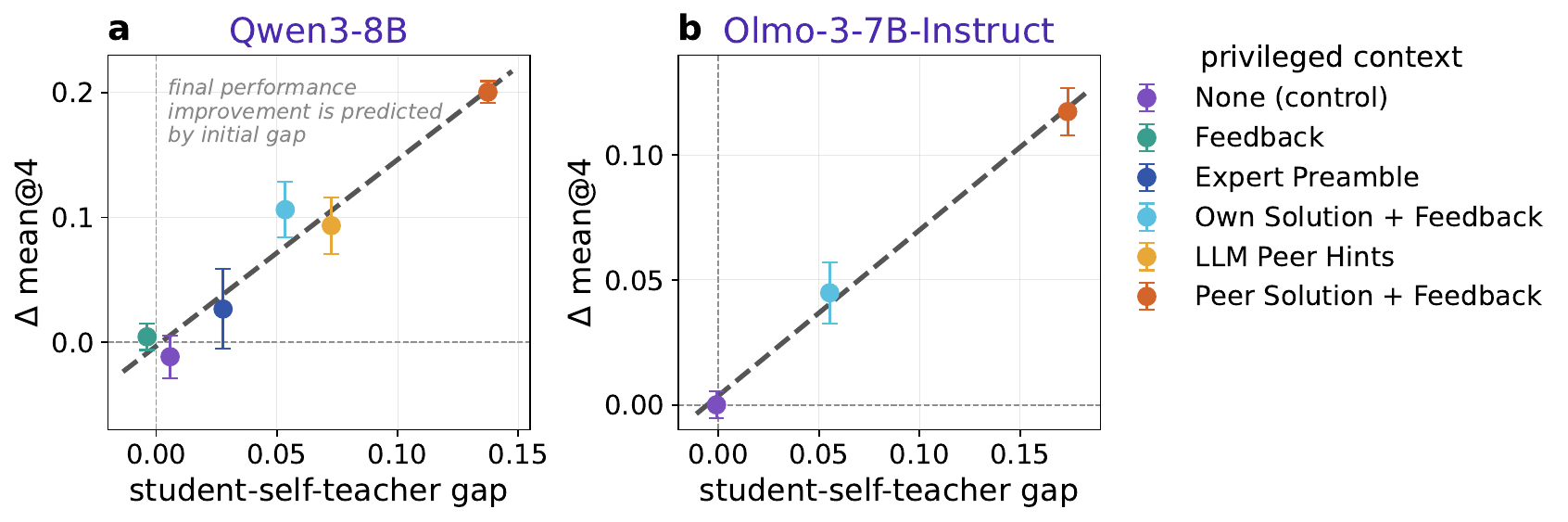}
\end{center}
    \caption{
    \textbf{Linear correlation between initial student–self-teacher gap and final student performance improvement}. \textbf{a)} Final student improvement versus the initial gap between the student and its self-generated teacher, for Qwen3-8B. Each dot represents the mean improvement for a given context configuration (see legend and Section~\ref{sec:methodology}), with error bars indicating one standard deviation across 3 random seeds. The black dashed line shows the ordinary least squares (OLS) linear fit (intercept~$= -0.003$, slope~$= 1.492$, $R^2 = 0.949$). The estimated Pearson and Spearman correlations between final student improvement and the initial gap are $0.974$ ($p = 0.001$) and $0.886$ ($p = 0.019$), respectively. \textbf{b)} Same as \textbf{a)}, for Olmo-3-7B-Instruct. OLS linear fit: intercept~$= 0.004$, slope~$= 0.663$, $R^2 = 0.996$. Estimated Pearson and Spearman correlations: $0.998$ ($p = 0.043$) and $1.0$ ($p = 0.0$), respectively.
    }
\label{fig:figure-1}
\end{figure*}

\section{Introduction}
Reinforcement learning with verifiable rewards (RLVR) has become a fundamental component of the training pipeline for large language models. Unlike supervised fine-tuning (SFT), RLVR allows the model to sample rollouts, explore on-policy behavior, and train on resulting feedback, leading to enhanced reasoning capabilities~\cite{jaech2024openai,guo2025deepseekr1, kimiteam2025kimik15}. Despite its success, RLVR is constrained by outcome-level rewards, which are often binary. This limits the use of richer feedback signals, such as interpreter messages in coding environments or detailed feedback from environment~interactions.

To address this limitation, several OPSD approaches have been proposed \cite{hubotter2026reinforcement,shenfeld2026self, zhao2026self, ye2026online, sang2026policy, ye2026policy}. Rather than relying on scalar rewards, these methods generate dense, logit-level credit assignment directly from arbitrary tokenized feedback, such as the environment and/or the model's own behavior. Despite promising results, OPSD comes with notable challenges: privileged context conditioning shifts the self-teacher's distribution in unpredictable ways, and careful stabilization and context design are required for the approach to work reliably \cite{zhao2026self,partridge2026tweet2038715548071325794, brown2026tweet2050038277454143918}. For OPSD to become a reliable post-training method, a principled methodology is needed to predict the effect of the variables driving its success. In this work, we provide one such methodology by investigating an important yet overlooked property of OPSD: the privileged context construction. Like standard on-policy distillation (OPD) \cite{agarwal2024policy,lu2025onpolicydistillation}, OPSD targets the teacher distribution, but unlike OPD, OPSD allows the teacher’s effective performance ceiling to be varied through the choice of the privileged context.

Crucially, because the teacher is the student with privileged context (the \emph{self-teacher}), the initial performance gap between the student and self-teacher (the \emph{student–self-teacher gap}) is cheap to measure before any training. We show empirically that this initial gap is a strong linear predictor of final performance improvement, providing a practical lever to (1) screen multiple privileged context constructions in advance, (2) understand training limits to avoid wasted post-training runs, and most importantly, (3) enable reliable use of world feedback during post-training.

To do so, we focus on coding as a canonical environment with rich feedback (e.g., runtime errors, failed unit tests), using LiveCodeBench as our benchmark \cite{jain2024livecodebench}. We post-train open-source models following standard OPSD techniques and investigate the student–self-teacher predictive law through an extensive set of experiments. In sum, our work makes the following contributions:
\begin{enumerate}
\item We identify a strong predictive law between the initial student–self-teacher gap and final student performance improvement, enabling early performance estimation without costly full training runs.
\item We demonstrate that this relationship generalizes across different types of privileged information and model families. 
\item We show that the relationship holds with model size.
\end{enumerate}
Together, these findings offer a lightweight method for estimating the effectiveness of OPSD configurations during post-training, elevating world feedback as a reliable component of LLM post-training.

\section{Preliminaries}
\paragraph{Problem formulation.}
We adopt the OPSD formulation of \citet{hubotter2026reinforcement}. The student policy $\pi_\theta$ samples rollouts $y \sim \pi_\theta(\cdot \mid x)$ for prompts $x$ drawn from a prompt distribution $\mathcal{D}$. The self-teacher $\pi_{\bar\theta}(\cdot \mid x, c)$ shares its weights with the student via an exponential moving average (EMA), where $\bar\theta$ slowly drifts toward student parameters $\theta$ during training, and is conditioned on privileged context $c$ that contains information unavailable to the student (e.g., world feedback, reference snippets). Although the self-teacher and student have (almost) the same weights, $c$ shifts the teacher's predictive distribution toward stronger behavior through in-context learning, providing dense supervision without the need for a stronger base model. The OPSD objective is the per-token reverse KL divergence between the student and self-teacher under student rollouts:
\newsavebox{\opsdbox}
\newcommand{\opsdbody}[2]{%
  \mathcal{L}_{\text{OPSD}}(\theta; c) = #1 \mathbb{E}_{\substack{y \sim \pi_\theta(\cdot|x)\\x \sim \mathcal{D}}}\bigg[ \frac{1}{|y|}\sum_t \mathrm{KL}\Big( \pi_\theta(\cdot \mid y_{<t}, x) #2 \mathrm{stopgrad}[\pi_{\bar\theta}(\cdot \mid y_{<t}, x, c)] \Big)\bigg],%
}

\savebox{\opsdbox}{$\displaystyle \opsdbody{}{,\,}$}

\ifdim\wd\opsdbox>\linewidth
  \begin{align}
  \opsdbody{&}{,\nonumber\\ &\,}
  \end{align}
\else
  \begin{equation}
  \usebox{\opsdbox}
  \end{equation}
\fi
where $\mathrm{stopgrad}$ blocks the gradients through the teacher.

\section{Experiments}\label{sec:experiments}
In our experiments, we perform a series of post-training experiments with OPSD on two model families: Qwen3~\cite{yang2025qwen3} and Olmo 3 \cite{olmo2026olmo3}. We investigate the design of privileged context $c$ and the resulting performance correlation between the student and self-teacher models. For each model, we start from a pre-trained checkpoint and post-train using OPSD for 50 steps on LiveCodeBench, varying the context $c$ across runs. We construct six distinct contexts~$c$ by combining different information sources: the original prompt, the model's own rollouts, peer rollouts from the group, the rollout's success verdict, and external world feedback. The full set of context types is described in~Table~\ref{tab:context-types}.
For each context, we report two metrics: (1) initial student–self-teacher gap: the difference between the self-teacher and student accuracy on the validation set before training, using train-time decoding parameters; and~(2)~final student improvement ($\Delta$ mean@4): the gain in student accuracy on the validation set between start and end of training, using validation-time decoding parameters. More details are provided in Section~\ref{sec:methodology}.

\begin{figure}[ht]
\begin{center}
\includegraphics[width=.45\linewidth]{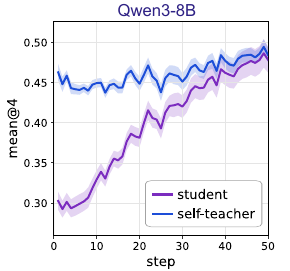}
\end{center}
    \caption{
    \textbf{The student performance gradually approaches the self-teacher performance throughout training.} During post-training of Qwen3-8B on LiveCodeBench with the {\ttfamily Peer Solution + Feedback} privileged context, the student validation accuracy (purple) steadily converges to the self-teacher level (blue). We report the mean (solid line) and standard deviation (shaded region) over 3 seeds.
    }
\label{fig:figure-1-training-dynamics}
\end{figure}

\subsection{Initial student–self-teacher gap predicts final student accuracy improvements.}\label{subsection:experiments-main}
We first test how the initial gap between the student and the self-teacher correlates with the student's final performance across different types of privileged contextual information. We use Qwen3-8B and Olmo-3-7B-Instruct as starting checkpoints. We observe a striking correlation between the initial student–self-teacher gap and final performance improvement for both model families (Figure~\ref{fig:figure-1}).
Notably, this linear relationship holds consistently across all tested types of privileged context, despite the qualitatively different content they convey (e.g., environment feedback, peer solutions, or general advice from the self-teacher).
Furthermore, we examine the training dynamics and find that the student's evaluation performance steadily converges to the self-teacher's level over time (Figure~\ref{fig:figure-1-training-dynamics}).
We also note that we see a rise in self-teacher capabilities over time similar to \citet{hubotter2026reinforcement}.
Finally, we conduct a leave-one-out cross-validation analysis to test the predictive capability of this empirical rule. The rule generalizes well across all folds, with $R^2 = 0.949$ and a root mean squared error (RMSE) of $0.016$ for Qwen3-8B and $R^2 = 0.996$ and RMSE $= 0.003$ for Olmo-3-7B-Instruct. This supports a strong and general underlying connection between the student–self-teacher gap and final student performance improvement.

\begin{takeaway}
The initial student–self-teacher gap strongly and linearly predicts final model performance improvement across privileged context types and model families.
\end{takeaway}

\begin{figure}[ht]
\begin{center}
\includegraphics[width=.5\linewidth]{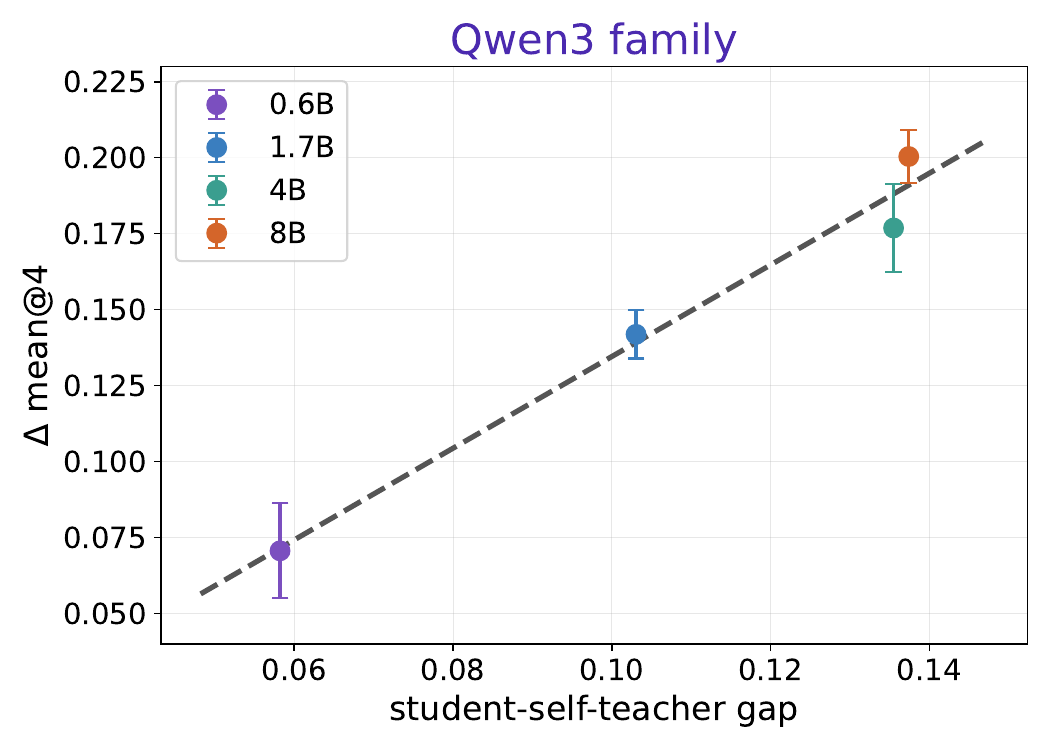}
\end{center}
    \caption{
    \textbf{The predictive law is preserved at scale}. Final student performance improvement vs. initial student–self-teacher gap across model sizes in the Qwen3 family. The initial gap and final performance improvement are computed as in Figure \ref{fig:figure-1}. Each dot represents the mean final performance for a given model size, with error bars showing standard deviation over 3 seeds. OLS linear fit: intercept~$= -0.016$, slope~$= 1.508$, $R^2 = 0.977$ with estimated Pearson and Spearman correlations: $0.988\ (p = 0.012)$ and $1.0\ (p = 0.0)$.
    }
\label{fig:figure-2}
\vspace{-0.1cm}
\end{figure}

\subsection{The predictive law holds with model size.}\label{subsection:experiments-scaling}
Next, we examine whether this predictive law holds across model scale. We use Qwen3 models of 4 sizes (0.6B, 1.7B, 4B, 8B) and measure the initial student–self-teacher gap and final performance improvement for a fixed privileged context type {\ttfamily Peer Solution + Feedback}. The linear relationship between the student–self-teacher gap and final student performance  improvement holds consistently across all model scales (Figure~\ref{fig:figure-2}), suggesting it is a reliable, scale-invariant predictor of final performance gain. Crucially, this not only means that final student performance can be estimated early in training, but also that performance gains at larger model sizes can be accurately predicted using this linear relationship.

\begin{takeaway}
The student–self-teacher gap is a scale-invariant, linear predictor of final performance improvement, enabling accurate prediction of performance gain across model sizes.
\end{takeaway}

\section{Discussion}
We introduce a predictive law for estimating final student performance from the initial student–self-teacher gap in OPSD. Empirical validation across Qwen3 and Olmo 3 model families on LiveCodeBench reveals a robust linear correlation between the initial gap and final performance improvement, consistent across model families and scales. This relationship has two practical implications. First, it enables screening multiple privileged information configurations without costly post-training runs. Second, it reliably predicts the benefit of incorporating world feedback during post-training, which is especially valuable in rich environment interaction settings. Together, these allow practitioners to estimate the performance ceiling for different privileged contexts prior to any training.

Furthermore, OPSD may serve as a practical alternative to OPD in settings where useful world feedback is available to bootstrap learning. OPD performance is often non-monotonic as stronger teachers do not always improve the student: reasoning trajectories can diverge \cite{li2026rethinkingonpolicydistillationlarge}, and reverse-KL objectives may destabilize high-capacity students~\cite{ko2026scalingreasoningefficientlyrelaxed}. In practice, recipes restrict teacher selection, e.g., to matched-size teachers or the same model family \cite{lu2025onpolicydistillation, ko2026scalingreasoningefficientlyrelaxed}. As a result, no predictive rules exist for OPD that forecast final performance without running experiments \cite{song2026survey}. In contrast, OPSD leverages the intrinsic consistency between student and self-teacher by construction, which we hypothesize explains our clean predictive law.

Our work opens several avenues for future research. First, it would be valuable to further study how different forms of privileged information affect final student performance and where they fall along the empirical predictive relationship. Methods that improve performance via in-context learning are a promising direction, including few-shot learning \cite{brown2020languagemodelsfewshotlearners}, retrieval-augmented generation \cite{lewis2021retrievalaugmentedgenerationknowledgeintensivenlp}, prompt evolution \cite{agrawal2026gepareflectivepromptevolution, khattab2023dspycompilingdeclarativelanguage}, and context evolution \cite{zhang2026agenticcontextengineeringevolving}. Second, scaling the law to larger models is an important next step to assess its generality. Finally, comparing the predicted gains from OPSD across different and richer environments would further establish its practical utility.

\section{Methodology}\label{sec:methodology}

\paragraph{Self-Teacher Construction Configurations.}
We study six privileged contexts $c$ that span qualitatively distinct information sources (Table~\ref{tab:context-types}). Notably, the configurations differ in \emph{kind}, not merely in strength, which allows the linear correlation in Figure~\ref{fig:figure-1} to test whether the scalar gap alone is a sufficient statistic for final performance regardless of the mechanism that produced it.

\paragraph{Models, Training, and Evaluation.}
We use Qwen3-\{0.6B, 1.7B, 4B, 8B\} \cite{yang2025qwen3} in non-thinking mode and Olmo-3-7B-Instruct \cite{olmo2026olmo3} as base policies. Each model is post-trained with OPSD for 50 gradient steps, batch size 32, and 8 rollouts per prompt; full hyperparameters are listed in Appendix~\ref{sec:experimental-details}. We evaluate on LiveCodeBench v6~\cite{jain2024livecodebench} (problems from February--May 2025) using mean@4 pass rate.

\begin{table}[t]
\centering
\caption{The six self-teacher constructions studied. Additional details in the Supplementary Materials (Table \ref{table:privileged-information}).}
\label{tab:context-types}
\begin{tabular}{ll}
\toprule
\textbf{Self-Teacher Construction} \\
\midrule
{\ttfamily None (control)} \\
{\ttfamily Expert Preamble} \\
{\ttfamily Feedback} \\
{\ttfamily LLM Peer Hints} \\
{\ttfamily Own Solution + Feedback} \\
{\ttfamily Peer Solution + Feedback} \\
\bottomrule
\end{tabular}
\end{table}

\section{Related Work}
\paragraph{RL with LLMs.}
Large-scale RL has significantly improved LLM performance on general reasoning tasks via  RLVR methods such as GRPO \cite{shao2024deepseekmath}. A complementary line of research seeks finer-grained credit assignment by training dedicated process reward models (PRMs) that score individual reasoning steps rather than final answers \cite{setlur2025scaling}. While RLVR approaches rely on outcome-level feedback and PRMs introduce significant overhead from a separate model, OPSD sidesteps both limitations by using the model as its own supervisor, requiring no external PRM.

\paragraph{On-policy distillation.}
When a capable teacher model is available, distillation offers a compelling alternative to standard SFT \cite{hinton2015distilling}. OPD represents a strong implementation of distillation because it avoids the exposure-bias gap of evaluating the student under the teacher's state distribution, and it is cheaper to sample from \cite{yang2025qwen3,lu2025onpolicydistillation}. By construction, OPD is bounded by the teacher distribution in the limit. By contrast, OPSD is bounded by (1) the performance gain from additional context, which can be adjusted dynamically during training, and (2) the self-teacher's capabilities, which can improve over time and raise the ceiling accordingly.

\paragraph{On-policy self-distillation.}
Our work is fundamentally related to OPSD approaches. In SDFT \cite{shenfeld2026self}, the self-teacher is constructed with expert demonstrations, allowing the model to distill task-relevant knowledge from in-context examples into its base parameters. In RLTF-SD \cite{song2026expanding}, the authors use textual feedback from annotators or users for the self-teacher, enabling the model to internalize corrective critiques. In ERL \cite{shi2026experiential}, the self-teacher is constructed with a structured reflection derived from environmental feedback on a failed first attempt, guiding a refined second attempt whose success is then consolidated into the policy. In OPCD \cite{ye2026policy}, historical solution traces or optimized system prompts are the self-teacher's privileged context, training the student to internalize experiential knowledge. In OPSD \cite{zhao2026self}, the self-teacher is constructed with verified correct reasoning traces. Finally, in SDPO \cite{hubotter2026reinforcement}, the self-teacher is conditioned on rich textual feedback from verifiable environments and correct model behavior. While these methods differ in their context configurations, they all share the same core approach: distilling in-context learning capabilities with privileged context into the student model.

\clearpage
\bibliographystyle{unsrtnat}
\bibliography{bib}

\clearpage
\appendix

\section{Experimental Details}\label{sec:experimental-details}
The experiments provided in Section~\ref{sec:experiments} are all performed on LiveCodeBench v6 \cite{jain2024livecodebench} (problems from February–May 2025). We build on the SDPO\footnote{\,\textcolor{TufaBlue}{\nolinkurl{https://github.com/lasgroup/SDPO}}} \cite{hubotter2026reinforcement} and VeRL\footnote{\,\textcolor{TufaBlue}{\nolinkurl{https://github.com/verl-project/verl}}} \cite{Sheng_2025} codebases. We use the same set of hyperparameters for all our experiments, which are provided in Table~\ref{table:hyperparameters}.

\begin{table}[h]
\centering
\caption{Hyperparameters.}
\begin{tabular}{ll}
\toprule
\textbf{Parameter} & \textbf{Value} \\
\midrule
\multicolumn{2}{l}{\textit{Model and data}} \\
Base models & Qwen3, Olmo3 \\
Dataset & LiveCodeBench v6 \\
\midrule
\multicolumn{2}{l}{\textit{Training}} \\
Train batch size (prompts) & 32 \\
Mini-batch size & 1 \\
Optimizer & AdamW \\
Learning rate & $1\times10^{-6}$ \\
LR schedule & constant \\
Weight decay & 0.01 \\
Adam $(\beta_1, \beta_2)$ & $(0.9,\ 0.999)$ \\
Gradient clip (max norm) & 1.0 \\
\midrule
\multicolumn{2}{l}{\textit{Sampling --- training rollouts}} \\
Rollouts per prompt ($n$) & 8 \\
Temperature & 1.0 \\
Top-$p$ & 1.0 \\
Max prompt length & 2048 \\
Max response length & 32{,}768 \\
Thinking mode & disabled \\
\midrule
\multicolumn{2}{l}{\textit{Sampling --- validation}} \\
Samples per prompt ($n$) & 4 \\
Temperature & 0.6 \\
Top-$p$ & 0.95 \\
Max prompt length & 2048 \\
Max response length & 32{,}768 \\
Thinking mode & disabled \\
\midrule
\multicolumn{2}{l}{\textit{SDPO}} \\
Divergence & Reverse-KL \\
Teacher EMA rate & 0.01 \\
Distillation top-$k$ & 20 \\
\bottomrule
\end{tabular}
\label{table:hyperparameters}
\end{table}
Below, we provide more information on the specific experiments we ran to produce the figures in the main text.
\begin{enumerate}
    \item For the experiments in Subsection~\ref{subsection:experiments-main}, we run OPSD for 50 steps on Qwen3-8B for 3 seeds and Olmo-3-7B-Instruct for 2 seeds while varying the context according to the self-teacher constructions in Table~\ref{table:privileged-information}. We ran additional experiments with more context constructions, where the same law holds well, but omit them because they lack sufficient seed coverage.
    \item For the experiments in Subsection~\ref{subsection:experiments-scaling}, we run OPSD for 50 steps on Qwen3-\{0.6B, 1.7B, 4B, 8B\} for 3 seeds while fixing the privileged context type {\ttfamily Peer Solution + Feedback}. We note that we did initial experiments on Qwen3-14B; however, we believe that for larger models we might have to tune hyperparameters separately and hence do not include these inconclusive results here.
\end{enumerate}

\clearpage

\begin{table}[h]
\centering
\caption{Six self-teacher constructions: \texttt{\{prompt\}} is the original problem; \texttt{\{sol\}} is a correct solution (peer's, the rollout's own, or LLM-extracted hints, depending on the row); \texttt{\{fb\}} is environment feedback. Bracketed sections are conditional per sample.}
\label{tab:context-templates}
\small
\renewcommand{\arraystretch}{1.2}
\begin{tabular}{l p{0.55\linewidth}}
\toprule
\textbf{Self-Teacher Construction} & \textbf{Template} \\
\midrule
{\ttfamily None (control)} & {\ttfamily \{prompt\}} \\
\addlinespace
{\ttfamily Expert Preamble} &
{\ttfamily [expert preamble: DS/algo checklist; complexity-vs-input-size table; pattern rules; edge cases; 9-step methodology]\newline ---\newline
\{prompt\}\newline Think step by step first, then write the solution.} \\
\addlinespace
{\ttfamily Feedback} &
{\ttfamily \{prompt\}\newline The following is feedback from your unsuccessful earlier attempt:\newline \{fb\}\newline Correctly solve the original
question.} \\
\addlinespace
{\ttfamily LLM Peer Hints} &
{\ttfamily \{prompt\}\newline Correct solution:\newline \{sol\}\newline Correctly solve the original question.} \\
\addlinespace
{\ttfamily Own Solution + Feedback} &
{\ttfamily \{prompt\}\newline [Correct solution: \{sol\}]\quad(if rollout correct)\newline [Feedback: \{fb\}]\quad(else)\newline Correctly solve the
original question.} \\
\addlinespace
{\ttfamily Peer Solution + Feedback} &
{\ttfamily \{prompt\}\newline Correct solution:\newline \{sol\}\newline The following is feedback from your unsuccessful earlier attempt:\newline
\{fb\}\newline Correctly solve the original question.} \\
\bottomrule
\end{tabular}
\label{table:privileged-information}
\end{table}

\paragraph{Computational Resources.}
All experiments ran on a cluster with NVIDIA B200 GPUs. We estimate each run took around 20h, leading to a total runtime of approximately 720 B200-hours, plus roughly 240 B200-hours for debugging and failed runs.

\end{document}